\def\paperlanguage{}
\pdfoutput=1 

\documentclass[letterpaper, 10 pt, conference]{ieeeconf}  

\usepackage{bm}
\usepackage{cite}
\include{preamble}

\IEEEoverridecommandlockouts                              

\title{\LARGE \textbf
  {
    \switchlanguage%
    {%
      Reflex-based Motion Strategy of Musculoskeletal Humanoids under Environmental Contact Using Muscle Relaxation Control
    }%
    {%
      筋弛緩制御を用いた環境接触下における動作戦略
    }%
  }
}

\author{Kento Kawaharazuka$^{1}$, Kei Tsuzuki$^{1}$, Moritaka Onitsuka$^{1}$, Yuya Koga$^{1}$\\Yusuke Omura$^{1}$, Yuki Asano$^{1}$, Kei Okada$^{1}$, Koji Kawasaki$^{2}$,  and Masayuki Inaba$^{1}$
  \thanks{$^{1}$ The authors are with the Department of Mechano-Informatics, Graduate School of Information Science and Technology, The University of Tokyo, 7-3-1 Hongo, Bunkyo-ku, Tokyo, 113-8656, Japan.
    {\texttt\small [kawaharazuka, tsuzuki, onitsuka, koga, oomura, asano, k-okada, inaba]@jsk.t.u-tokyo.ac.jp}
  }
  \thanks{$^{2}$ The author is associated with TOYOTA MOTOR CORPORATION.
    {\texttt\small koji\_kawasaki@mail.toyota.co.jp}
  }
}
\begin{document}

\maketitle
\thispagestyle{empty}
\pagestyle{empty}

\begin{abstract}
  \switchlanguage%
  {%
    The musculoskeletal humanoid can move well under environmental contact thanks to its body softness.
    However, there are few studies that actively make use of the environment to rest its flexible musculoskeletal body.
    Also, its complex musculoskeletal structure is difficult to modelize and high internal muscle tension sometimes occurs.
    To solve these problems, we develop a muscle relaxation control which can minimize the muscle tension by actively using the environment and inhibit useless internal muscle tension.
    We apply this control to some basic movements, the motion of resting the arms on the desk, and handle operation, and verify its effectiveness.
  }%
  {%
    筋骨格ヒューマノイドはその柔軟な身体構造ゆえに環境接触を伴う動作を得意とする.
    しかし, それらの環境を積極的に用いて力を分散させたり, 身体を休ませたりする動作制御は難しい.
    また, その複雑でモデル化困難な身体構造により, 高い筋内力が発生してしまう問題がある.
    これらの問題を解決するために, 環境を積極的に用いて必要な筋張力を最低限に保ち, かつ無駄な筋内力を抑制することができる, 筋弛緩制御を開発する.
    本制御を, 基本動作, 机を利用して身体を休める動作, ハンドル回しに応用し, 有効性を確かめた.
  }%
\end{abstract}

\section{INTRODUCTION}\label{sec:introduction}
\begin{figure}[t]
  \centering
  \includegraphics[width=0.9\columnwidth]{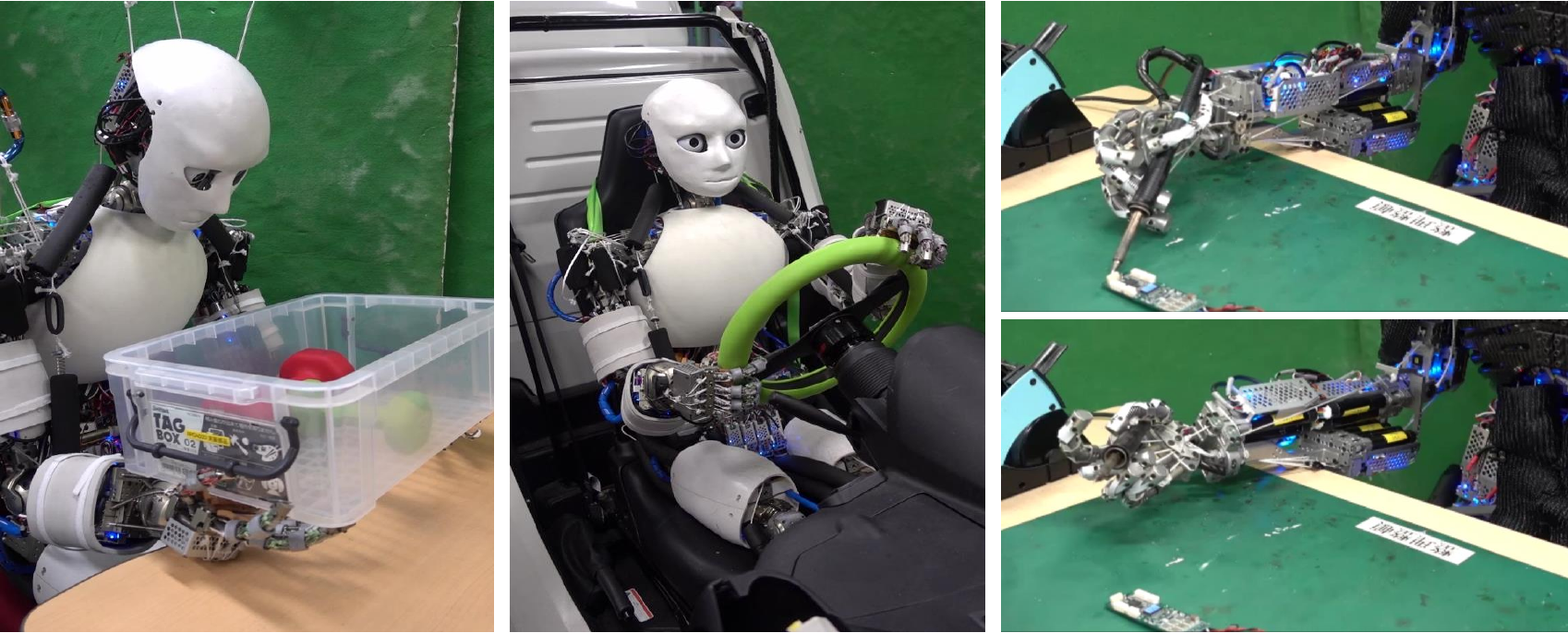}
  \caption{Motions with environmental contact by musculoskeletal humanoids.}
  \label{figure:motivation}
  \vspace{-3.0ex}
\end{figure}

\switchlanguage%
{%
  The musculoskeletal humanoid \cite{nakanishi2013design, wittmeier2013toward, jantsch2013anthrob, asano2016kengoro} has many biomimetic benefits such as muscle redundancy, variable stiffness control using nonlinear elastic elements, ball joints without extreme points, and the under-actuated fingers and spine.
  Also, thanks to its under-actuation, flexibility, and impact resistance, the musculoskeletal humanoid is suitable for motions with environmental contact as shown in \figref{figure:motivation}, compared to the ordinary axis-driven humanoid \cite{hirai1998asimo, hirukawa2004hrp}.
  However, it is difficult for the musculoskeletal humanoid to continuously execute tasks with environmental contact, while inhibiting useless internal forces.
  There are two main reasons.

  First, there is a large error between the actual robot and its geometric model, because the complex musculoskeletal structure is difficult to modelize.
  High internal muscle tension sometimes occurs due to the inaccurate antagonistic relationship, the muscle temperature rises rapidly, and the robot may break down.
  To solve this problem, various methods have been proposed.
  Mizuuchi, et al. have constructed a neural network representing the relationship between joint angles and muscle lengths (joint-muscle mapping, JMM) from motion capture data, and moved the robot using the network \cite{mizuuchi2006acquisition}.
  Kawaharazuka, et al. have constructed JMM online using vision information, and updated JMM to decrease internal muscle tension \cite{kawaharazuka2018online}.
  Also, they have controlled the internal muscle tension by including muscle tension in JMM \cite{kawaharazuka2018bodyimage, kawaharazuka2019longtime}.
  Although these methods can solve the problem of model error to a certain degree, muscle hysteresis cannot be modelized, and so several reflex-based control methods, have been developed.
  Asano, et al. have developed a load sharing method of muscles by using the actual sensor information \cite{asano2013loadsharing}.
  Kawaharazuka, et al. have developed an antagonist inhibition control to permit model error by loosening antagonist muscles \cite{kawaharazuka2017antagonist}.
  These methods are executed rapidly, automatically, and without consciousness, compared to other feedforward controls.
  However, the former is required to manually decide muscles to share the load, and the latter is sensitive to the difference between the target and current joint angles, so they are not practical.
}%
{%
  筋骨格ヒューマノイド\cite{nakanishi2013design, wittmeier2013toward, jantsch2013anthrob, asano2016kengoro}は, 筋の冗長性や非線形弾性要素による可変剛性制御, 特異点のない球関節や劣駆動な背骨・指等, 生物規範型の様々な利点を有する.
  また, その劣駆動性・柔軟で冗長な腱・衝撃に強い球関節等の特徴から, 通常の軸駆動型ヒューマノイド\cite{hirai1998asimo, hirukawa2004hrp}に比べ, 環境接触下における動作実現に適している(\figref{figure:motivation}).
  しかし, 筋骨格ヒューマノイドにおいて内力を抑えて継続的に環境接触行動を行うことは難しい.
  これには, 2つの問題が起因していると考える.
  ただし, 筋骨格ヒューマノイドは摩擦等の観点から筋張力制御が難しく\cite{kawamura2016jointspace}, 本研究では筋長制御を前提とする.

  1つめに, 筋骨格ヒューマノイドはその複雑な身体構造ゆえにモデル化が難しく, 実機と幾何モデルの間に大きな誤差が生じる点である.
  そのため, 筋の拮抗関係の誤差により内力がたまり, 筋温度が急上昇したり, ロボットが破損する可能性がある.
  これに関しては様々な解決方法が提案されている.
  水内らはモーションキャプチャから得た関節角度と実機の筋長を対応付けたニューラルネットワークを構築し, それを元に制御指令を計算している\cite{mizuuchi2006acquisition}.
  河原塚らは, 視覚から得た関節角度と筋長を対応付けたニューラルネットワークを獲得すると同時に, 内力を減らすようにネットワークをオンラインで更新する手法を開発している\cite{kawaharazuka2018online}.
  また, 関節角度と筋長だけでなく, 筋張力もネットワークに加えることで, 内力を制御し可変剛性制御を実現している\cite{kawaharazuka2018bodyimage}.
  上記の手法は内力をある程度解消することが可能であるが, ヒステリシス等のモデル化困難な項は依然として残り, それらを解決するために反射型の手法も開発されている.
  浅野らは実機センサデータから筋張力を分配することで, 筋の負荷を和らげる手法を開発している\cite{asano2013loadsharing}.
  河原塚らは主動筋に対して拮抗筋を緩めることでモデル誤差を許容する拮抗筋抑制制御を開発している\cite{kawaharazuka2017antagonist}.
  しかし, 前者は負荷を分散する筋群を手動で決める必要があり, 後者は指令関節角度と関節角度推定値との差に敏感であるため静的動作において上手く機能しない等, 様々な問題を抱える.
}%

\switchlanguage%
{%
  Second, because it is difficult for the musculoskeletal humanoid to move accurately, internal force between the robot and environment sometimes occurs.
  Thus it is difficult for the robot to actively make use of the environment and rest its body.
  To solve this problem, adding muscle elasticity is effective \cite{osada2010addon, nakanishi2012absorption}, and the errors of recognition and movement can be permitted flexibly.
  On the other hand, long-time internal force is a burden to the muscles, and useless internal force should be omitted.
  Also, there are few studies that actively make use of the environment.

  In this study, as a solution of these two problems, we propose a muscle relaxation control (MRC).
  This is a simple reflex-based control to relax unnecessary muscles without influencing the current posture.
  This method can inhibit internal muscle tension, and the robot can move continuously.
  Also, MRC can inhibit internal force between the robot and environment, and the robot can rest its body by making use of the environment.
  In this study, we use muscle length-based controls because muscle tension-based controls of the musculoskeletal humanoid are difficult due to muscle friction \cite{kawamura2016jointspace}.
}%
{%
  2つめに, 筋骨格ヒューマノイドは身体の正確な位置決めが難しく, 環境との接触により内力が発生したり, 逆に環境を上手く利用したりすることが難しい点である.
  前者に関しては, 筋骨格ヒューマノイドの腱が持つ弾性, そして非線形弾性が効果を示す\cite{osada2010addon, nakanishi2012absorption}.
  環境認識や動作に誤差があっても, それを柔軟に許容可能である.
  しかし同時に, 長時間の内力は筋に対して負荷となり, 必要のない内力は除去すべきである.
  また, 環境を積極的に利用した動作研究は乏しい.

  本研究では, これら2つの問題を解決する手法として, 筋弛緩制御を提案する.
  これは, 姿勢に影響を及ぼさない範囲で筋を弛緩させていくというシンプルな制御である.
  本研究を用いることで, 1つめの問題である拮抗関係における筋内力を抑え継続的な動作が可能となると同時に, 2つめの問題である環境との接触による内力の抑制, そして環境を使った身体を休める動作等が可能となる.
  \secref{sec:musculoskeletal-structure}では, 本研究で用いる筋骨格ヒューマノイドの基本構造について述べる.
  \secref{sec:proposed}では, 本研究の提案である筋弛緩制御とそれを含む全体システム, また, その特徴について述べる.
  \secref{sec:experiments}では, 基本動作・重量物体把持・環境を利用して身体を休める動作・ハンドル動作を扱う.
  最後に, \secref{sec:conclusion}において, これら手法と実験について結論を述べる.
}%

\begin{figure}[t]
  \centering
  \includegraphics[width=0.6\columnwidth]{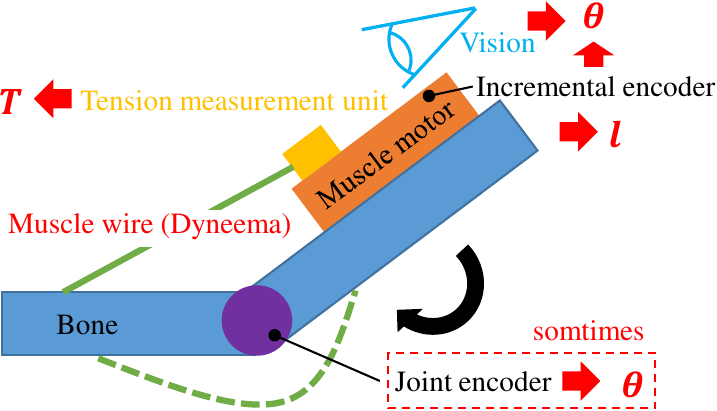}
  \caption{The basic structure of musculoskeletal humanoids handled in this study.}
  \label{figure:musculoskeletal-structure}
  \vspace{-3.0ex}
\end{figure}

\section{Basic Musculoskeletal Structure} \label{sec:musculoskeletal-structure}
\switchlanguage%
{%
  We show the basic musculoskeletal structure handled in this study in \figref{figure:musculoskeletal-structure}.
  Muscles are wound by pulleys using electrical motors, and antagonistically arranged around joints.
  Muscle length $\bm{l}$, tension $\bm{T}$, and temperature $\bm{C}$ can be measured.
  The abrasion resistant synthetic fiber Dyneema is used for the muscle wire, and it is slightly elastic.
  Some robots have a nonlinear elastic element at the end of each muscle to enable variable stiffness control.
  Also, the joint angle $\bm{\theta}$ can be measured in some robots, and can be estimated using the muscle length and vision information even if it cannot be directly measured \cite{kawaharazuka2018online}.

  In this study, we represent $M$ and $N$ as the number of muscles and joints, respectively.
  $\bm{l}$, $\bm{T}$, $\bm{C}$ are $M$ dimensional vectors, and $\bm{\theta}$ is a $N$ dimensional vector.
  $\{\bm{l}, \bm{T}, \bm{C}\}^{i}$ represents an element of the vector regarding $i^{th}$ muscle.
}%
{%
  本研究で扱う筋骨格ヒューマノイドの基本的な構造を\figref{figure:musculoskeletal-humanoid}の左下図に示す.
  関節に対して筋が拮抗に冗長配置されており, 筋の筋長と筋張力, 筋温度を測定することができる.
  筋アクチュエータは空気圧等ではなく, 電気モータを使用している.
  筋は摩擦に強い化学繊維であるダイニーマを使用しており, 筋自体が多少伸びる.
  ロボットによっては筋の末端に可変剛性制御を可能とする非線形弾性要素が配置されていることがある.
  また, ロボットによっては関節角度が測定できるように工夫されており, 関節角度が測定できない場合も筋長変化と視覚を用いることで実機関節角度を推定することができる\cite{kawaharazuka2018online}.

  本研究で使用する筋骨格ヒューマノイドMusashi\cite{kawaharazuka2019musashi} (\figref{figure:musculoskeletal-humanoid}の左上)は, 筋に非線形弾性要素を, 関節には学習型制御模索のために関節角度センサを有している.
  本研究では主に, 肩の3自由度, 肘の2自由度を用いて実験を行うこととし, 関節角度は$\bm{\theta}=(\theta_{S-r}, \theta_{S-p}, \theta_{S-y}, \theta_{E-p}, \theta_{E-y})$のように表す($S$はshoulder, $E$はelbow, $rpy$はそれぞれroll, pitch, yawを表す).
  また, 肩と肘の5自由度は筋を10本含み, その筋配置は\figref{figure:musculoskeletal-humanoid}の右上のようになっている.
}%

\section{Muscle Relaxation Control} \label{sec:proposed}
\switchlanguage%
{%
  Muscle Relaxation Control (MRC) is a control to relax muscles in order from unnecessary ones, without changing the current posture.
  The musculoskeletal humanoid has agonist and antagonist muscles; agonist muscles mainly execute the current task and antagonist muscles follow them.
  Thus, MRC relaxes muscles in order from unnecessary antagonist ones, and then relaxes agonist muscles too.

  We will first explain the whole system including MRC, and then explain the detailed implementation of MRC and its characteristics.
}%
{%
  まず筋弛緩制御を含む全体のシステムについて説明する.
  その後, 本研究の提案である筋弛緩制御, そして筋弛緩制御の特徴について述べる.
}%

\subsection{The Whole System} \label{subsec:whole-system}
\begin{figure}[t]
  \centering
  \includegraphics[width=1.0\columnwidth]{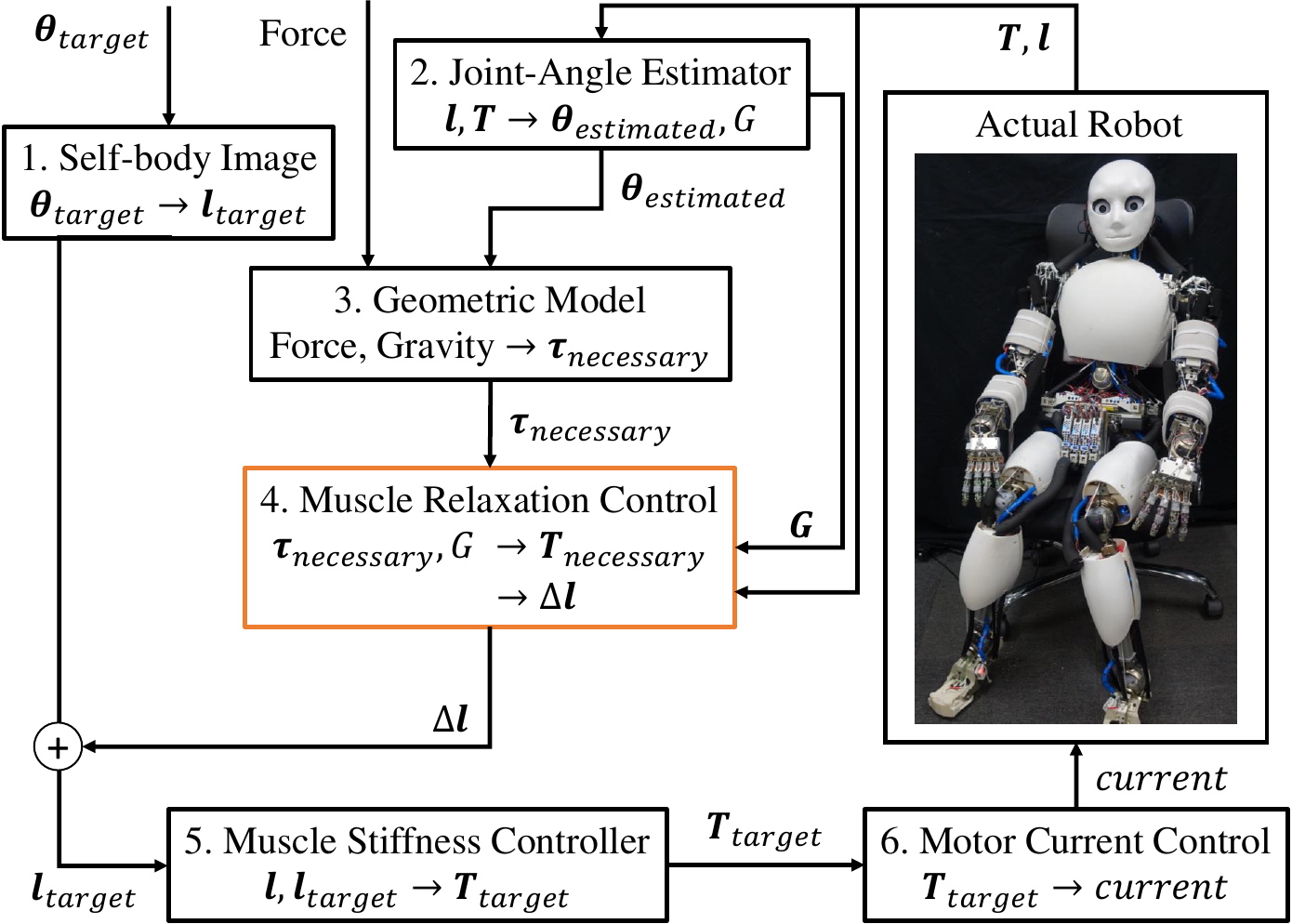}
  \caption{The whole system of this study.}
  \label{figure:whole-system}
  \vspace{-3.0ex}
\end{figure}

\switchlanguage%
{%
  The whole system including MRC is shown in \figref{figure:whole-system}.

  Regarding 1 (Self-body Image), the target joint angle $\bm{\theta}_{target}$ is converted into target muscle length $\bm{l}_{target}$ by using the self-body image \cite{kawaharazuka2018bodyimage}.
  By using the robot model updated using the actual robot sensor information, an accurate joint angle and internal muscle tension can be realized.
  However, there are problems of friction, hysteresis, etc., and we can improve them by reflex-based controls.

  Regarding 2 (Joint-Angle Estimator), the current joint angle $\bm{\theta}_{estimated}$ and the current muscle Jacobian $G$ are estimated using the self-body image \cite{kawaharazuka2018bodyimage} with the current muscle tension $\bm{T}$ and muscle length $\bm{l}$.

  Regarding 3 (Geometric Model), the current necessary joint torque $\bm{\tau}_{necessary}$ is calculated from the necessary force $\bm{F}$ and $\bm{\theta}_{estimated}$ using a geometric model, which has information of joint position, link length, and link weight.

  Regarding 4 (Muscle Relaxation Control), from the calculated muscle relaxation value $\Delta\bm{l}$, which will be explained in the next section, the target muscle length is updated as shown below.
  \begin{align}
    \bm{l}_{target} \gets \bm{l}_{target} + \Delta\bm{l}
  \end{align}

  Regarding 5 (Muscle Stiffness Controller), the target muscle tension $\bm{T}_{target}$ is calculated by muscle stiffness control \cite{shirai2011stiffness} as shown below,
  \begin{align}
    \bm{T}_{target} = \bm{T}_{bias} + \textrm{max}(\bm{0}, K_{stiff}(\bm{l}-\bm{l}_{target})) \label{eq:muscle-stiffness-control}
  \end{align}
  where $\bm{T}_{bias}$ is a bias value of muscle tension, and $K_{stiff}$ is a muscle stiffness gain.

  Regarding 6 (Motor Current Control), the target current is calculated from $\bm{T}_{target}$ and is sent to the actual robot.

  In these procedures, the frequencies of 2 and 3 are 40 Hz, those of 1, 4, and 5 are 125 Hz, and that of 6 is 1000 Hz.
}%
{%
  全体のシステムは\figref{figure:whole-system}のようになっている.
  一つ一つのコンポーネントについて説明する.

  まず, 1では自己身体像\cite{kawaharazuka2018bodyimage}を用いて指令関節角度$\bm{\theta}_{target}$を指令筋長$\bm{l}_{target}$に変換する.
  これまでは幾何モデルを用いて関節角度を筋長に変換していたが, 実機センサデータをもとに更新された身体モデルを利用することで, より正確な関節角度・適切な内力を実現することができる.
  しかし, 筋の摩擦・ヒステリシス等の問題により, 反射型の制御による改善の余地が見られる.

  次に, 2では自己身体像\cite{kawaharazuka2018bodyimage}を用いた関節角度推定・筋長ヤコビアン$G$の導出を行う.
  これは, 拡張カルマンフィルタを用いて筋長変化から関節角度を推定する手法\cite{ookubo2015learning}における多項式近似を自己身体像に置き換えて行うことができる\cite{kawaharazuka2018bodyimage}.

  次に, 3では現在のタスクにおける必要な力や関節角度推定値$\bm{\theta}_{estimated}$, 重力等から, 現在必要なトルク$\bm{\tau}_{necessary}$を計算する.

  次章で説明する4の筋弛緩制御において得られた筋弛緩項$\Delta\bm{l}$から, $\bm{l}_{target}$を以下のように更新する.
  \begin{align}
    \bm{l}_{target} = \bm{l}_{target} + \Delta\bm{l}
  \end{align}

  5では, 筋剛性制御\cite{shirai2011stiffness}により以下のように指令筋張力$\bm{T}_{target}$が計算される.
  \begin{align}
    \bm{T}_{target} = \bm{T}_{bias} + \textrm{max}(\bm{0}, K_{stiff}(\bm{l}-\bm{l}_{target})) \label{eq:muscle-stiffness-control}
  \end{align}
  ここで, $\bm{T}_{bias}$は筋剛性制御のバイアス項, $K_{stiff}$は筋剛性制御の筋剛性係数である.

  6において, 最終的に電流値が計算され, 実機に送られる.

  この中で, 2, 3は20 msec周期, 1, 4, 5は8 msec周期, 6は 1 msec周期で行われている.
}%

\begin{figure*}[t]
  \centering
  \includegraphics[width=1.5\columnwidth]{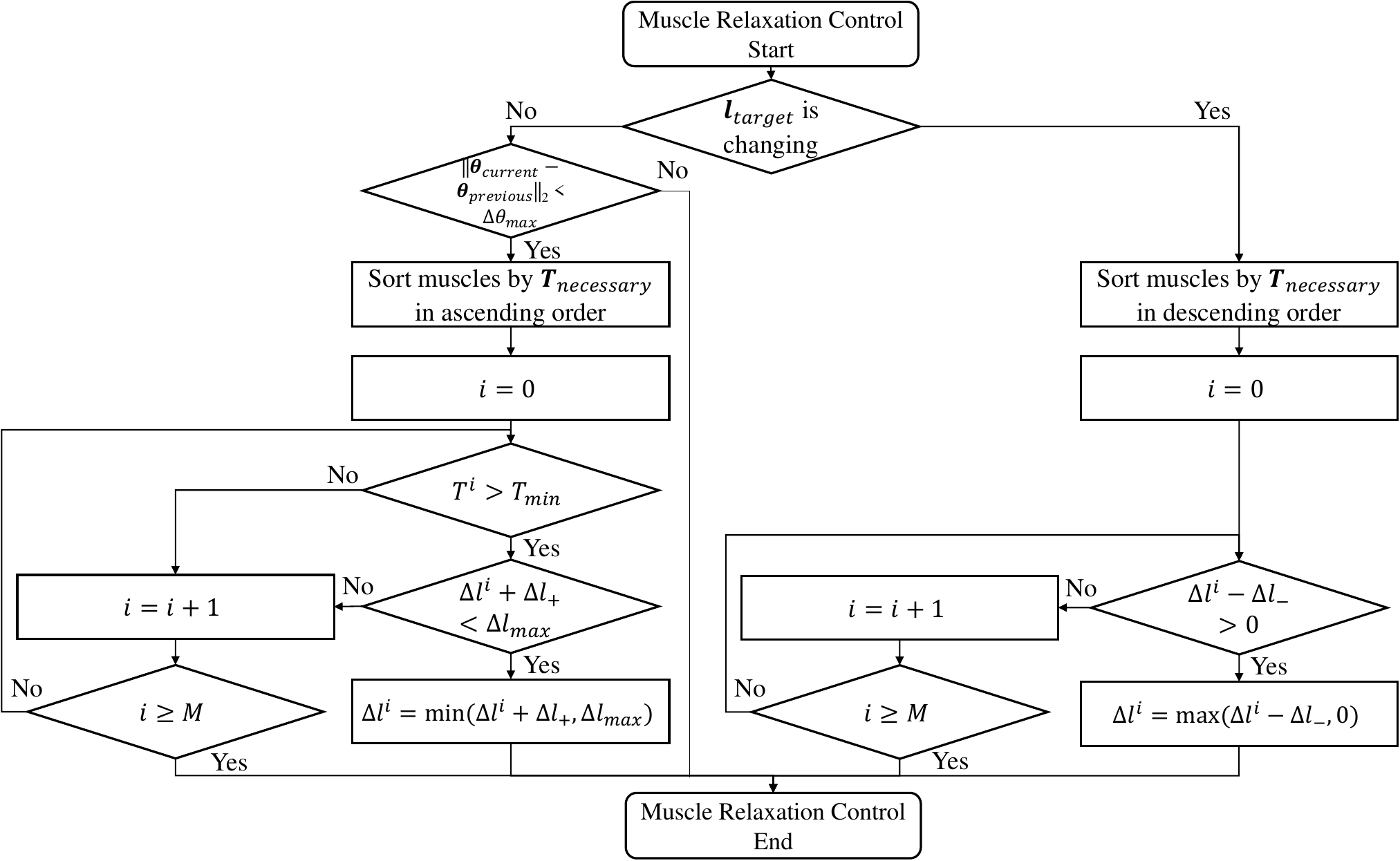}
  \caption{Flow chart of muscle relaxation control during one time step.}
  \label{figure:relaxation-control}
  \vspace{-1.0ex}
\end{figure*}

\subsection{Muscle Relaxation Control} \label{subsec:relaxation-control}
\switchlanguage%
{%
  MRC has a simple control structure.
  First, the current necessary muscle tension $\bm{T}_{necessary}$ is calculated by quadratic programming from $G$ and $\bm{\tau}_{necessary}$, as shown below.
  We set the calculated $\bm{x}$ as $\bm{T}_{necessary}$.
  \begin{align}
    \underset{\bm{x}}{\textrm{minimize}}&\;\;\;\;\;\;\;\bm{x}^{T}W_{1}\bm{x} + (G^{T}\bm{x}+\bm{\tau}_{nec})^{T}W_{2}(G^{T}\bm{x}+\bm{\tau}_{nec})\\
    \textrm{subject to}&\;\;\;\;\;\;\;\;\;\;\;\;\;\;\;\;\;\;\;\;\;\;\;\;\;\;\;\;\;\bm{x} \geq \bm{T}_{min}
  \end{align}
  where $W_{1}$ and $W_{2}$ are weights, $\bm{T}_{min}$ is a minimum muscle tension, and $\bm{T}_{nec}$ is the abbreviation of $\bm{T}_{necessary}$.
  We set $W_{1}$ and $W_{2}$ as $I\times1.0E-5$ and $I$, respectively ($I$ means the identity matrix).
  In this procedure, although an equality constraint is normally set to $\bm{\tau}_{necessary}=-G^{T}\bm{x}$, we applied the above formulation to permit the error of $G$, in order to avoid the case in which a solution cannot be found.

  Second, the entire process shown in \figref{figure:relaxation-control} is executed during each time step.
  We first set a current muscle relaxation value $\Delta{\bm{l}}$ to add to $\bm{l}_{target}$ as $\bm{0}$.
  $\Delta{l}_{\{+, -\}}$ represents a constant value to increase or decrease $\Delta\bm{l}$ by at one time step, and $\Delta{l}_{max}$ represents a maximum value of $\Delta{l}$.
  First, muscles are sorted by $\bm{T}_{necessary}$ in ascending order, and the $i^{th}$ muscle is checked in order from the most unnecessary for the task, with small ${T}^{i}_{necessary}$.
  If the current muscle tension $T^{i}$ is lower than $T_{min}$, the next muscle $i+1$ is checked.
  In the case of $T^{i}>T_{min}$, the muscle relaxation value $\Delta{l}^{i}$ is examined.
  If $\Delta{l}^{i}+\Delta{l}_{+}\geq\Delta{l}_{max}$, the next muscle $i+1$ is checked.
  If $\Delta{l}^{i}+\Delta{l}_{+}<\Delta{l}_{max}$, $\Delta{l}^{i}$ is updated as $\textrm{min}(\Delta{l}^{i} + \Delta{l}_{+}, \Delta{l}_{max})$.
  If $\Delta{l}^{i}$ can be updated, this procedure is finished, and if not, the above procedures are repeated until the last muscle.

  Although this is the basic procedure of MRC, there are several conditions.
  First, the muscle relaxation process is executed only at a static state (the target muscle length $\bm{l}_{target}$ is not changing).
  While $\bm{l}_{target}$ is changing, muscles are sorted by $\bm{T}_{necessary}$ in descending order, and $\Delta{l}^{i}$ ($\Delta{l}^{i} > 0$) with the largest $T^{i}_{necessary}$ is updated as $\Delta{l}^{i} = \textrm{max}(\Delta{l}^{i}-\Delta{l}_{-}, 0)$ at each time step.
  As with the static state, if $\Delta{l}^{i}$ can be updated, this procedure is finished, and if not, the above procedures are repeated until the last muscle.
  Also, MRC is stopped when $||\bm{\theta}_{current}-\bm{\theta}_{previous}||_{2}\geq\Delta{\theta}_{max}$ ($\bm{\theta}_{previous}$ is the joint angle when $\bm{l}_{target}$ stops changing, $\bm{\theta}_{current}$ is the current joint angle, and $||\cdot||_2$ means L2 norm).
  This is because MRC should not largely influence the body posture.
  As $\bm{\theta}_{current}$, we can use $\bm{\theta}_{estimated}$, the value of joint angle sensors in the case the robot has them, or the estimated joint angle calculated from muscle length and vision information \cite{kawaharazuka2018online}.

  In this study, we set $\bm{T}_{min}=30$ [N], $\Delta{l}_{-}=\Delta{l}_{+}=0.03$ [mm], $\Delta{l}_{max}=2.0$ [mm], and $\Delta{\theta}_{max}=0.1$ [rad].
}%
{%
  筋弛緩制御は非常に単純な制御構造をしている.

  まず, 2, 3から得られた$G, \bm{\tau}_{necessary}$から, 現在必要と考えられる筋張力$\bm{T}_{necessary}$を以下のような二次計画法を解くことで計算する.
  \begin{align}
    \textrm{minimize}&\;\;\;\;\;\;\;\;\;\;\;\;\;\;\;\;\;\;\;\;\;\;\;\;\;\;\;\;\;\bm{x}^{T}W_{1}\bm{x}\nonumber\\
    &\;\;\;\;\;+ (G^{T}\bm{x}+\bm{\tau}_{necessary})^{T}W_{2}(G^{T}\bm{x}+\bm{\tau}_{necessary})\\
    \textrm{subject to}&\;\;\;\;\;\;\;\;\;\;\;\;\;\;\;\;\;\;\;\;\;\;\;\;\;\;\;\;\;\bm{x} \geq \bm{T}_{min}\\
  \end{align}
  ここで, $W_{1}, W_{2}$は重みを表し, 本研究ではそれぞれ$I\times1.0E-5$, $I$とする($I$は単位行列を表す).
  計算された$\bm{x}$を$\bm{T}_{necessary}=\bm{x}$とする.
  これは, 通常は$\bm{\tau}_{necessary}=-G^{T}\bm{x}$とするのが一般的であるが, $G$の誤差から解が求まらない可能性があるため, より誤差を許容できるこの形を採用した.

  次に, $\bm{T}_{necessary}$の値から筋を昇順にソートし, $\bm{T}_{necessary}$の小さい, つまり本タスクに必要のない筋$i$から順に見ていく.
  本手順を\figref{figure:relaxation-control}に示す.
  現在の実機の筋張力センサ値$T^{i}$が$T^{i}<T_{min}$の場合は次の筋$i+1$へ移動する.
  もし$T^{i}>T_{min}$の場合は, $\Delta{l}^{i}$の値を確認する.
  もし$\Delta{l}^{i}+\Delta{l}_{+}>\Delta{l}_{max}$ならば, 次の筋$i+1$へ移動する.
  $\Delta{l}^{i}+\Delta{l}_{+}<\Delta{l}_{max}$ならば, $\Delta{l}^{i} = \textrm{min}(\Delta{l}^{i} + \Delta{l}_{+}, \Delta{l}_{max}$とし, $\Delta{l}^{i}$を更新する.
  一度でも$\Delta{l}^{i}$の更新に成功すればループを抜け, 成功しない場合は最後の筋まで上記の工程を繰り返す.

  これが基本的な筋弛緩制御の動きであるが, その作動に関してはいくつかの条件が入る.
  まず, 筋弛緩制御は筋長指令$\bm{l}_{target}$が変動しない, つまり静止状態においてのみ動作する.
  $\bm{l}_{target}$が変動している, つまり動作中に関しては, 逆に$\bm{T}_{necessary}$が高い筋から順に, $\Delta{l}^{i} = \textrm{max}(\Delta{l}^{i}-\Delta{l}_{-}, 0)$のように, $\Delta{l}^{i}$を元に戻していく.
  これも同様に, 一度でも$\Delta{l}^{i}$の更新に成功すればループを抜け, 成功しない場合は最後の筋まで上記の工程を繰り返す.
  また, $\bm{l}_{target}$が静止した瞬間の関節角度を$\bm{\theta}_{previous}$, 現在の関節角度を$\bm{\theta}_{current}$としたときに, $||\bm{\theta}_{current}-\bm{\theta}_{previous}||_{2}>\Delta{\theta}_{max}$となった場合には筋弛緩制御を停止する.
  これは, 筋の弛緩によって姿勢に大きな影響を及ぼさないようにするためである.
  $\bm{\theta}_{current}$は関節角度推定値$\bm{\theta}_{estimated}$を用いても良いし, \secref{sec:musculoskeletal-structure}にもあるように, 関節角度センサが備わっている場合はその値, 備わっていない場合は視覚と筋長変化から算出することができ\cite{kawaharazuka2018online}, それらを用いることもできる.

  本研究では, $\bm{T}_{min}=30$ [N], $\Delta{l}_{-}=\Delta{l}_{+}=0.03$ [mm], $\Delta{l}_{max}=2.0$ [mm], $\Delta{\theta}_{max}=0.1$ [rad]とする.
}%

\subsection{Characteristics of Muscle Relaxation Control} \label{subsec:relax-characteristics}
\switchlanguage%
{%
  By executing MRC, regarding ordinary movements, internal muscle tension decreases while relaxing antagonist muscles.
  When finishing the relaxation of antagonist muscles and starting the relaxation of agonist muscles, MRC stops because the current posture changes.
  Also, in the case of motions while constraining the body and environment, the posture does not change even if agonist muscles relax.
  Thus, not only antagonist muscles but also agonist muscles relax, and not only internal muscle tension but also internal force between the body and environment gradually decreases.

  While MRC can be applied to basic movements or motions with environmental contact, its applicable fields are limited by some conditions.
  MRC cannot be used with variable stiffness control, which changes joint stiffness using internal muscle tension, and impedance control.
  However, MRC can be applied to motions such as wiping a table, operating a handle, and leaning onto a shelf, and we should use MRC and other controls accordingly.
}%
{%
  筋弛緩制御は簡単に言うと, 姿勢を崩さない範囲内で, 必要のない筋から順に筋を弛緩させていくことに相当する.
  筋骨格ヒューマノイドは主動筋と拮抗筋を有し, その動作を主に遂行する筋が主動筋, それに追従する筋が拮抗筋である.
  つまり, なるべく必要のない拮抗筋から緩ませ, 最終的には主動筋すら緩んでいくことになる.

  筋弛緩制御を実行することで, 通常の動作においては, 拮抗筋が弛緩することで, 内力の高まりを抑えることができる.
  拮抗筋が弛緩し終わり, 主動筋が緩む際には, 姿勢の崩れが起きるため, 主動筋の弛緩は途中で止まる.
  また, 棚を掴む等の環境に身体を拘束させて行う動作の際は, 主動筋を弛緩させてもある程度までは姿勢に崩れが起きることはない.
  よって, 拮抗筋だけでなく主動筋も緩んでいき, 身体の内力だけでなく環境との接触の内力も小さくなっていく.

  基本的な動作や環境接触には広く使えるものの, 筋弛緩制御の適用範囲はいくつかの制約を受ける.
  筋弛緩制御は内力により剛性を制御する可変剛性制御とは併用することはできない.
  また, 動作しながら環境に力を発揮する場合は良いものの, 動作せず一定の力を継続的に環境に発揮したい場合等には向いていない.
  しかし同時に, 継続的な強い力を必要としないテーブル拭きや電車のつり革に掴まる動作, ハンドル操作や棚に掴まる動作等, 多くの環境接触動作に適用可能である.
}%

\begin{figure}[t]
  \centering
  \includegraphics[width=1.0\columnwidth]{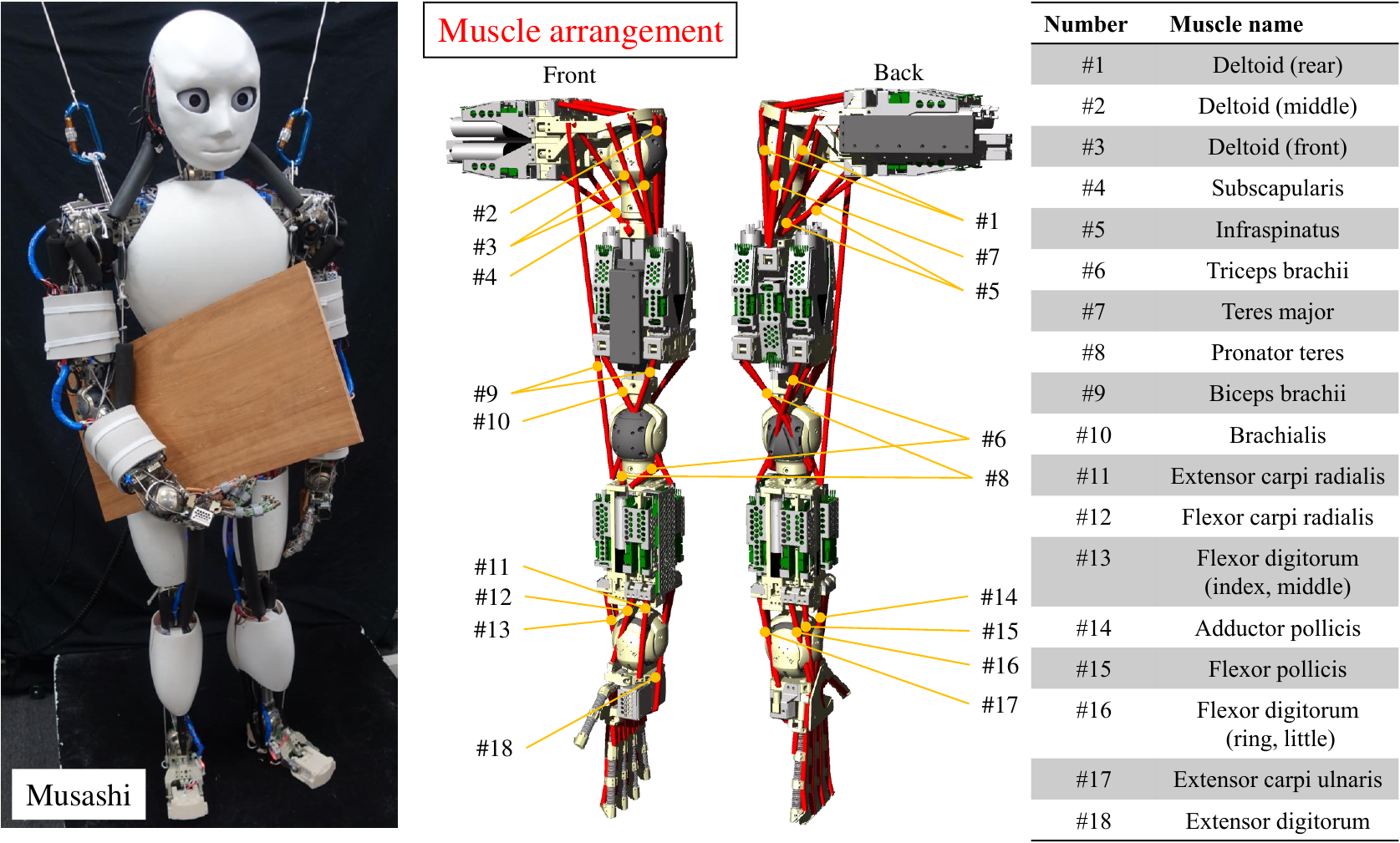}
  \vspace{-1.0ex}
  \caption{The musculoskeletal humanoid Musashi used for the following experiments.}
  \label{figure:musculoskeletal-humanoid}
  \vspace{-3.0ex}
\end{figure}

\section{Experiments} \label{sec:experiments}
\switchlanguage%
{%
  The musculoskeletal humanoid Musashi \cite{kawaharazuka2019musashi} (the left figure of \figref{figure:musculoskeletal-humanoid}) used for the following experiments is equipped with nonlinear elastic elements in the muscles, and it has joint angle sensors for the investigation of learning control systems.
  In the experiments, we mainly use the 3 degrees of freedom (DOFs) shoulder and 2 DOFs elbow for experiments, and we represent the joint angles as $\bm{\theta}=(\theta_{S-r}, \theta_{S-p}, \theta_{S-y}, \theta_{E-p}, \theta_{E-y})$ ($S$ means the shoulder, $E$ means the elbow, and $rpy$ means the roll, pitch, and yaw).
  These 5 DOFs include 10 muscles, $\#1$ -- $\#10$ in the muscle arrangement shown in the right figure of \figref{figure:musculoskeletal-humanoid}.
  Thus, $M=10$ and $N=5$.

  We executed several motions using the proposed MRC, and will discuss its function and effect in each situation.
  We show only the result of the left shoulder and elbow, even if we executed the experiment with both arms.
}%
{%
  本手法を用いて, いくつかの状況に応じた筋弛緩制御の動きと効果について実験・考察し, 有効的な使い方について説明する.
  単腕・双腕の実験についても左腕の肩と肘を合わせた結果のみ示し, \secref{sec:musculoskeletal-structure}に示したように関節角度は5自由度・筋は10本である.
}%

\subsection{Basic Movements}
\begin{figure}[t]
  \centering
  \includegraphics[width=0.9\columnwidth]{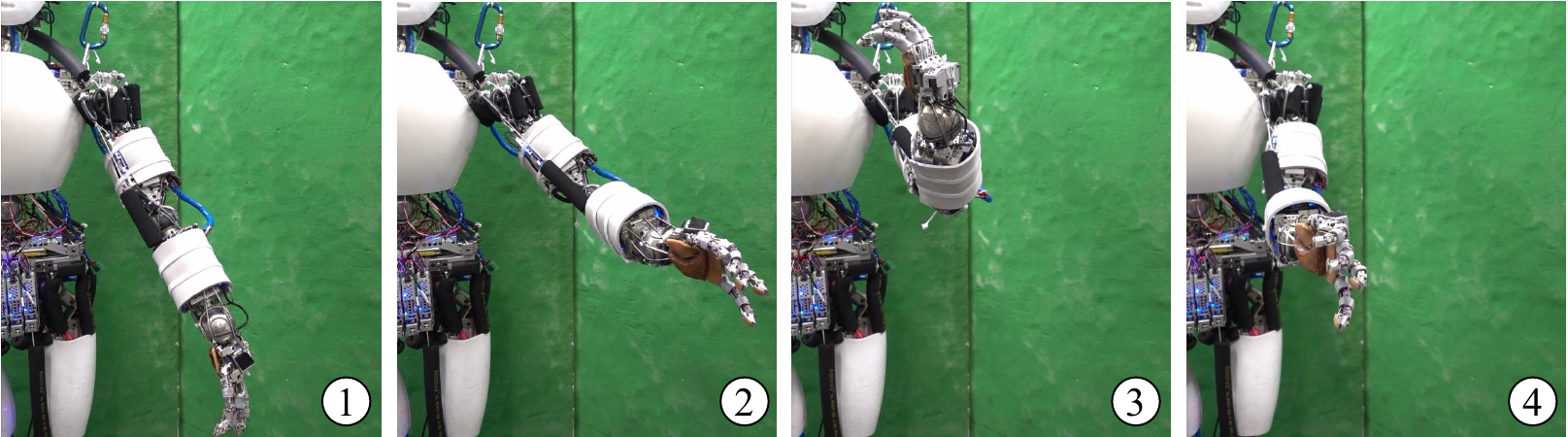}
  \caption{The experiment of basic movements with and without muscle relaxation control.}
  \label{figure:basic-experiment}
  \vspace{-1.0ex}
\end{figure}

\begin{figure}[t]
  \centering
  \includegraphics[width=1.0\columnwidth]{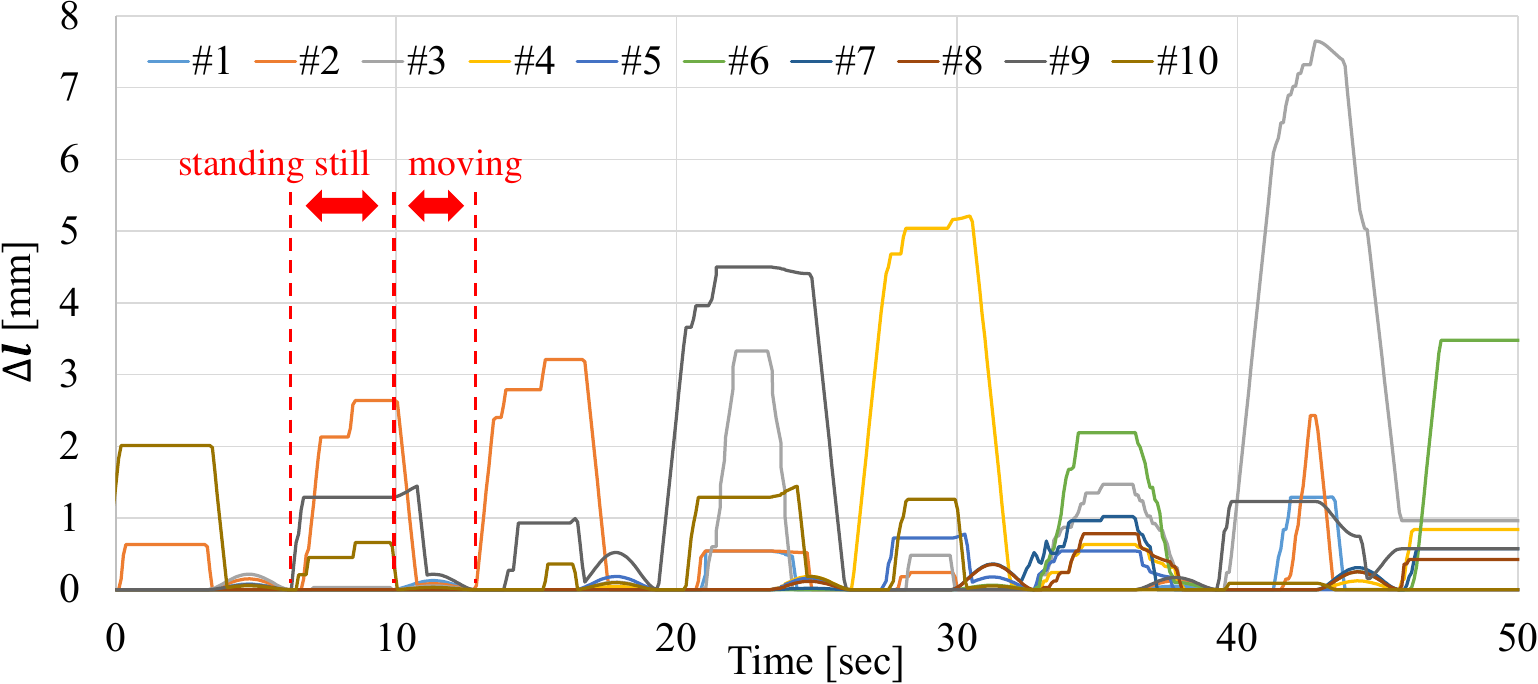}
  \caption{The transition of $\Delta{l}^{i}$ during basic movements with muscle relaxation control.}
  \label{figure:basic-length}
  \vspace{-1.0ex}
\end{figure}

\begin{figure}[t]
  \centering
  \includegraphics[width=0.9\columnwidth]{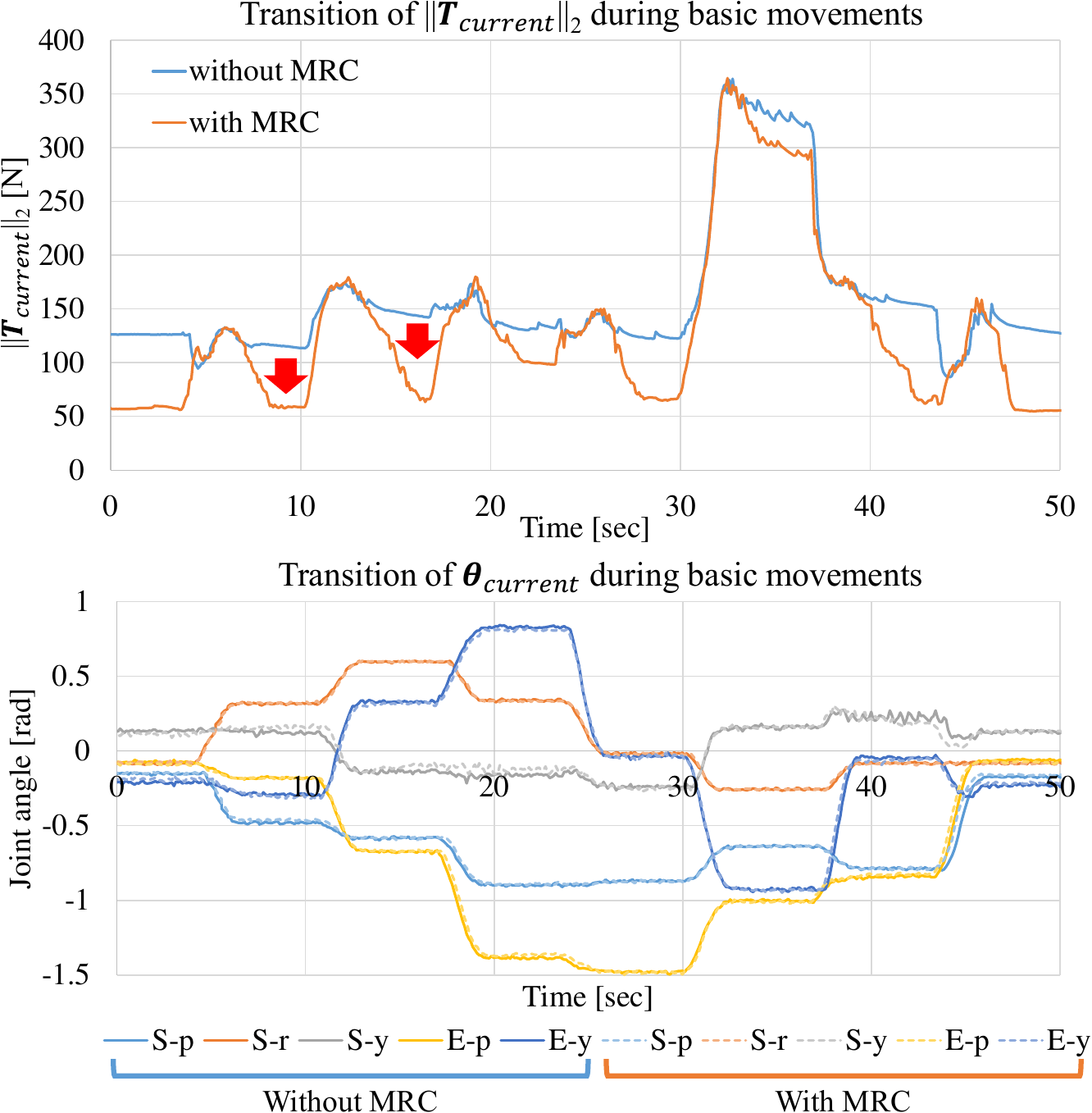}
  \caption{The transition of $||\bm{T}_{current}||_{2}$ and $\bm{\theta}_{current}$ during basic movements with and without muscle relaxation control.}
  \label{figure:basic-tension-angle}
\end{figure}

\switchlanguage%
{%
  We considered the effect of MRC when conducting basic movements.
  As shown in \figref{figure:basic-experiment}, we repeated movements of sending random joint angles over 3 sec and standing still for 3 sec.
  We examined the transition of $\Delta\bm{l}$, $\bm{\theta}_{current}$, and $||\bm{T}_{current}||_{2}$.

  First, we show the transition of $\Delta\bm{l}$ when using MRC in \figref{figure:basic-length}.
  $\Delta\bm{l}$ increases gradually by MRC in a static state, and it decreases when moving.
  The muscles whose $\Delta{l}^{i}$ changes largely are different depending on the movement, and MRC works on the muscles considered to be useless for the posture.

  Second, we show the transition of $||\bm{T}_{current}||_{2}$ and $\bm{\theta}_{current}$ during basic movements with and without MRC in \figref{figure:basic-tension-angle}.
  When using MRC, the muscle tension in a static state decreases largely compared to without MRC.
  This can be considered to be because useless internal muscle tension due to model error and friction is released.
  There is almost no difference of $\bm{\theta}_{current}$ between with and without MRC, and the tracking ability of joint angles is not changed by MRC.
  Thus, when conducting these basic movements, internal muscle tension can decrease largely without influencing the tasks.
}%
{%
  基本的な動作群を行い, その際の筋弛緩制御の動きについて考察する.
  \figref{figure:basic-experiment}のように関節角度をランダムに指定して3秒間で送り, 3秒間静止する動作を複数回行う.
  その際の$\Delta\bm{l}$, $\bm{\theta}_{current}$, $||\bm{T}_{current}||_{2}$の遷移を考察する.

  まず, 筋弛緩制御を入れた時の$\Delta\bm{l}$の遷移を\figref{figure:basic-length}に示す.
  静止した際にはMRCが働いて$\Delta\bm{l}$が上昇していき, 動作している際には$\Delta\bm{l}$が減少していく様子がわかる.
  $\Delta\bm{l}$が大きく変化する筋は動作によって異なり, 現在最も重要でないと考えられる筋に対してMRCが働いている.

  次に, 筋弛緩制御を入れた際と入れなかった際における$||\bm{T}_{current}||_{2}$の遷移の違いを\figref{figure:basic-tension}に示す.
  筋弛緩制御を入れた場合は, 入れなかった場合に比べて静止時に大きく筋張力が下がっていることがわかる.
  これは, モデル誤差や摩擦による無駄な筋内力が解放されたためであると考えられる.

  最後に, 筋弛緩制御を入れた際と入れなかった際における$\bm{\theta}_{current}$の遷移を\figref{figure:basic-angle}に示す.
  両者の違いはほとんどなく, 筋弛緩制御を入れることによって関節角度の追従性はほとんど変わらないことがわかる.
  よって, このような基本動作の際には, タスクに影響を与えずに, 筋内力を大きく減らすことが可能である.
}%

\subsection{Heavy Object Grasping}
\begin{figure}[t]
  \centering
  \includegraphics[width=0.9\columnwidth]{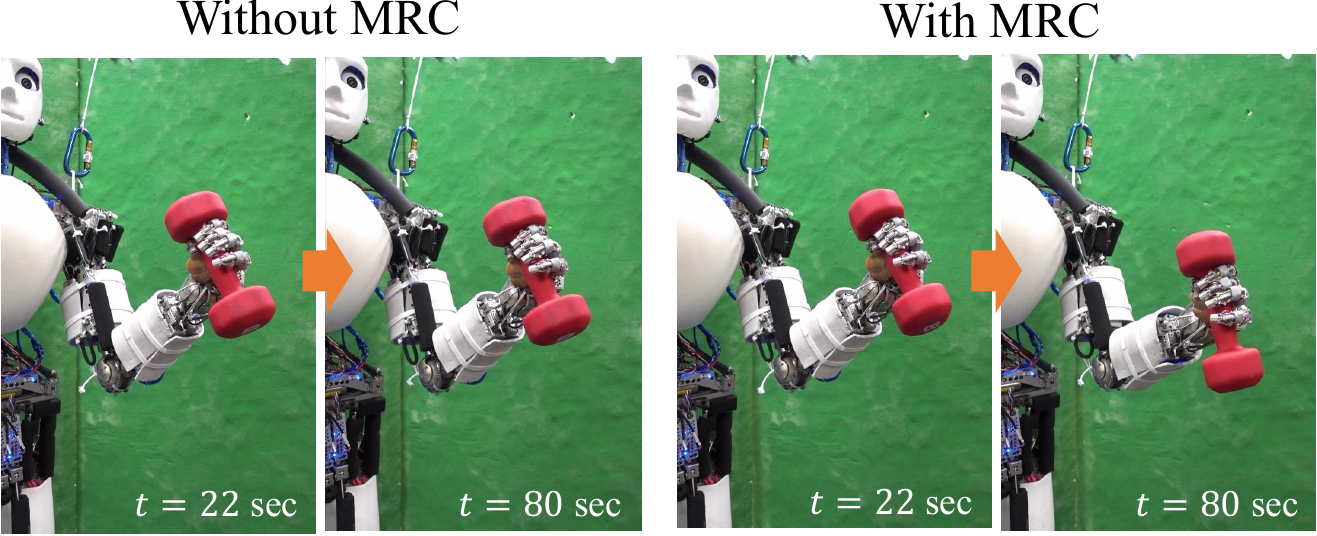}
  \caption{The experiment of grasping a heavy dumbbell.}
  \label{figure:dumbbell-experiment}
  \vspace{-3.0ex}
\end{figure}

\begin{figure}[t]
  \centering
  \includegraphics[width=0.9\columnwidth]{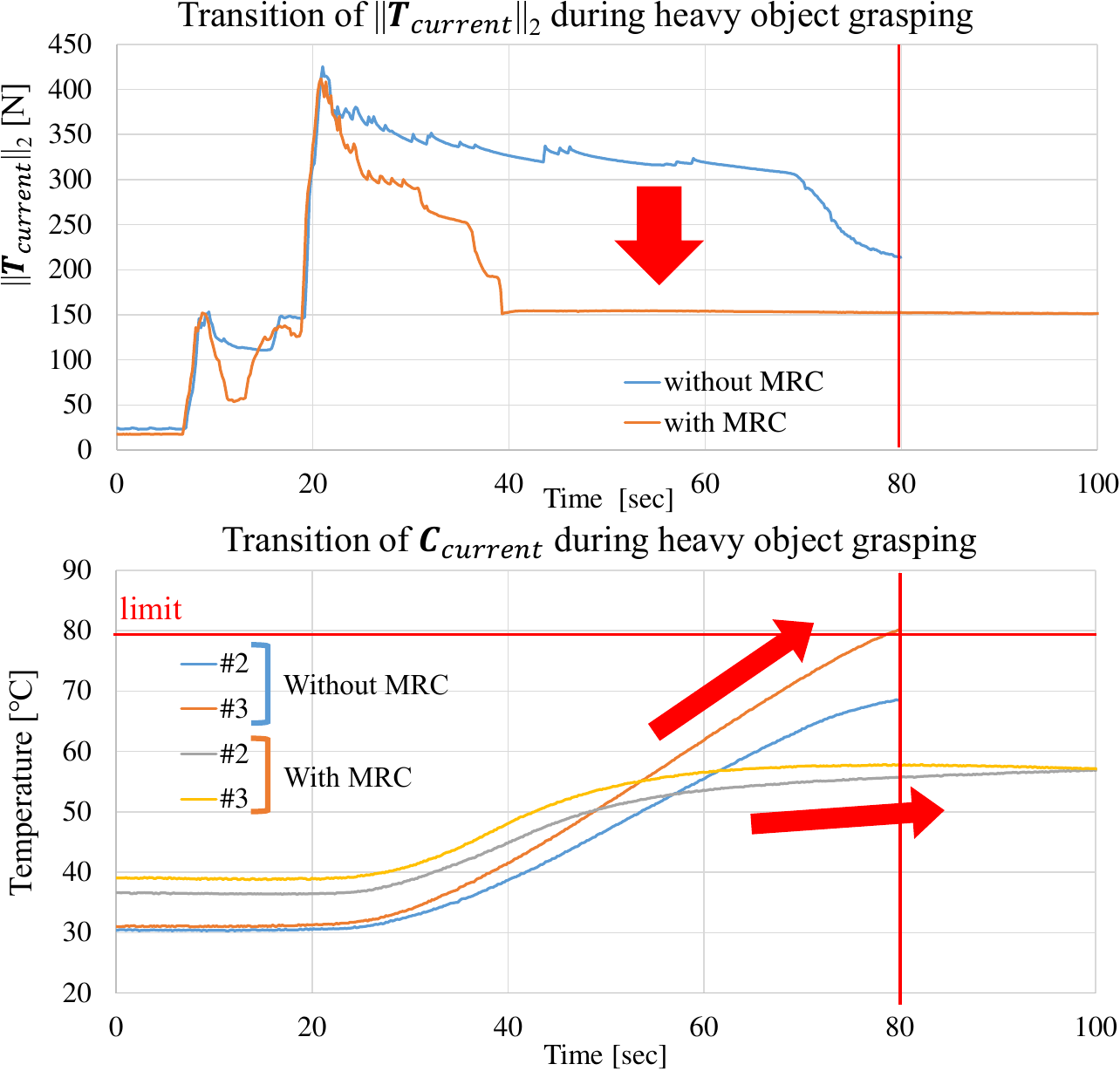}
  \caption{The transition of $||\bm{T}_{current}||_{2}$ and $\bm{C}_{current}$ while grasping a heavy dumbbell with and without muscle relaxation control.}
  \label{figure:dumbbell-tension-thermo}
  \vspace{-3.0ex}
\end{figure}

\switchlanguage%
{%
  We examined the function of MRC when grasping a heavy object like a dumbbell.
  As shown in \figref{figure:dumbbell-experiment}, Musashi lifted a 3.6 kg dumbbell and kept the posture.
  In this situation with and without MRC, we show the transition of $||\bm{T}_{current}||_{2}$ and 2 muscle temperatures $C_{current}$ with the highest muscle tensions in \figref{figure:dumbbell-tension-thermo}.
  Compared to without MRC, the muscle tension is inhibited largely with MRC.
  The reason that muscle tension gradually decreases even without MRC is because the safety mechanism of \cite{kawaharazuka2019longtime} begins to work when the muscle temperature exceeded 60 $^\circ$C.
  In the case with MRC, the muscle temperature does not exceed 60 $^\circ$C, and muscle tension is kept small constantly.
  In the case without MRC, the muscle temperature increases rapidly, and this experiment is stopped when it exceeded 80 $^\circ$C.

  While the robot can move continuously by inhibiting muscle tension with MRC, when a large force like in this experiment is applied, MRC permits the error of joint angles until the limit of $\Delta{\theta}_{max}$.
  Therefore we can see the joint angle error between with and without MRC at $t = 80$ sec in \figref{figure:dumbbell-experiment}.
  Although we can decrease the joint angle error by setting this $\Delta{\theta}_{max}$ to small value, the effect of MRC decreases and there exists a trade off of MRC.
  We must set $\Delta{\theta}_{max}$ appropriately depending on the task.
}%
{%
  ダンベルのような重量物体を把持した際の筋弛緩制御の働きについて考察する.
  \figref{figure:dumbbell-experiment}のように, 約3.6 kgのダンベルを持ち上げ, その状態で静止する.
  このときの筋弛緩制御を入れた場合と入れない場合における$||\bm{T}_{current}||_{2}$の遷移を\figref{figure:dumbbell-tension}に, 最も大きな筋張力を出していた2本の主動筋の温度遷移を\figref{figure:dumbbell-thermo}に示す.
  筋弛緩制御を入れない場合に比べて, 入れた場合は大きく筋張力を抑制できていることがわかる.
  ここで, 筋弛緩制御を入れない場合でも徐々に筋張力が下がっているのは, \cite{kawaharazuka2019longtime}における安全機構が, 温度が60度を超えたところで動作し始めているためである.
  筋弛緩制御を入れた場合は温度が60度を超えないため, ほぼ一定の低い筋張力を発揮している.
  筋弛緩制御を入れない場合は筋温度が急上昇し, 80度を超えたところで動作を終了している.

  筋弛緩制御を入れた方が筋張力を抑制し継続的に動作ができるものの, 大きな力が加わる場合は関節角度の誤差を$\Delta{\theta}_{max}$限界まで許容するため, 大きく関節角度誤差が発生していることがわかる.
  この$\Delta{\theta}_{max}$を小さく設定することで関節角度誤差は小さくできるものの, 筋張力抑制の効果は薄れてしまい, ここに本制御のトレードオフが存在していることがわかった.
}%

\subsection{Resting the Arms on the Desk}
\begin{figure}[t]
  \centering
  \includegraphics[width=0.8\columnwidth]{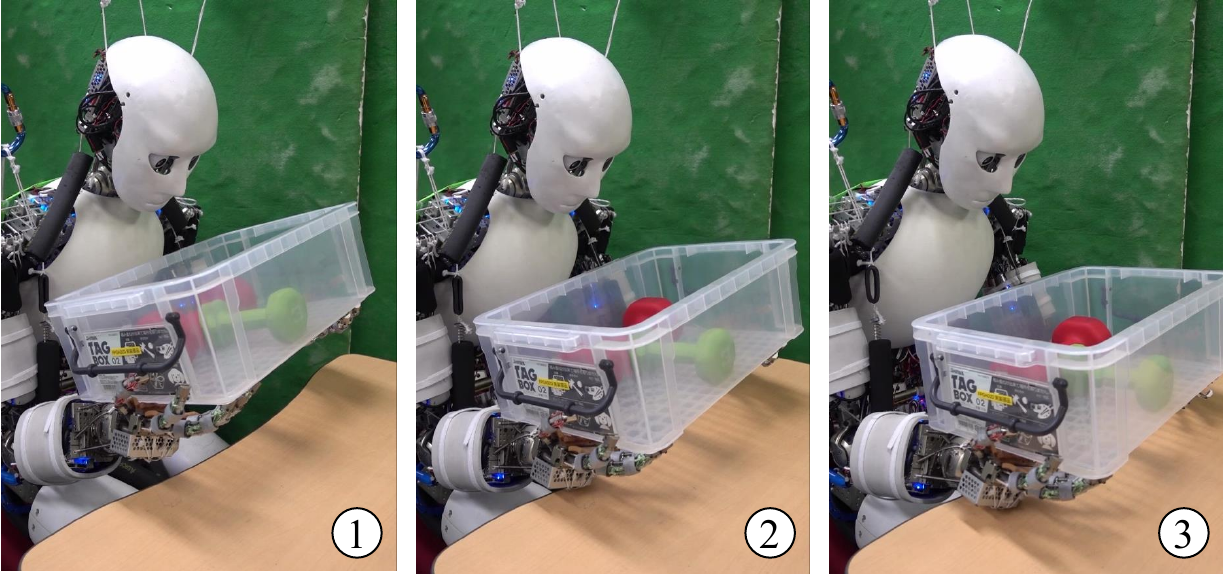}
  \caption{The experiment of resting the arms on the desk while carrying a heavy box.}
  \label{figure:box-experiment}
\end{figure}

\begin{figure}[t]
  \centering
  \includegraphics[width=0.8\columnwidth]{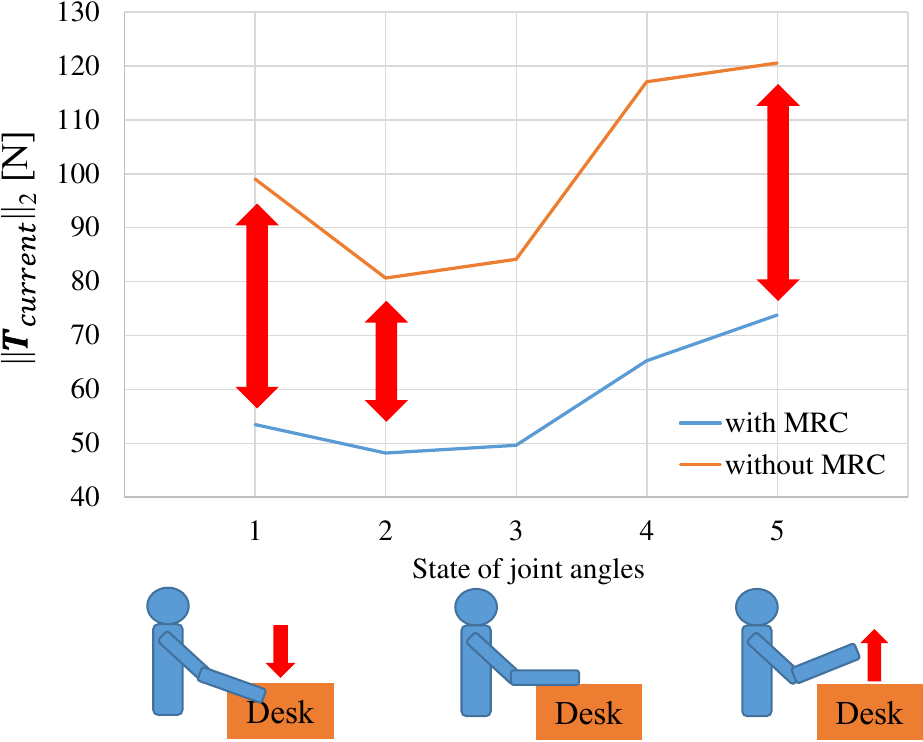}
  \caption{$||\bm{T}_{current}||_{2}$ while resting the arms on the desk with and without muscle relaxation control, regarding each joint angle state 1 -- 5.}
  \label{figure:box-tension}
  \vspace{-3.0ex}
\end{figure}

\switchlanguage%
{%
  We examined the function of MRC when resting the arms on the desk while carrying a heavy object.
  As shown in \figref{figure:box-experiment}, the robot held a 5 kg box, and rested the arms on the desk.
  In this situation, due to model or recognition error, the robot cannot always conduct the motion of leaning on the desk without large internal force between the robot and environment.
  In this experiment, we used only the joints of $S-p$ and $E-p$, we moved $(\theta_{S-p}, \theta_{E-p})$ to 5 states of 1: (-45, -80), 2: (-45, -85), 3: (-45, -90), 4: (-50, -90), and 5: (-50, -95) when resting the arms.
  As shown in the lower figure of \figref{figure:box-tension}, these states of joint angles show the transition of the robot arms from sinking into the environment to leaving the environment in order from 1 to 5.

  We show $||\bm{T}_{current}||_{2}$ at states 1 -- 5 with and without MRC in \figref{figure:box-tension}.
  Compared to without MRC, unnecessary antagonist muscle tension is inhibited and muscle tension decreases with MRC.
  Muscle tension at state 2 is the lowest with and without MRC.
  Muscle tension increases when sending the posture of sinking into the environment like in state 1.
  When leaving the environment like in state 5, muscle tension increases because the robot is not able to rest the arms.
  Also, when focusing on the difference of $||\bm{T}_{current}||_{2}$ between with and without MRC, the difference is the lowest in state 2 or 3, and the highest in state 1 or 5.
  This can be considered to be because the error between the arms and environment is absorbed with MRC, while whether the robot can rest the arms appropriately is sensitive to the posture without MRC.
  When sending the posture of sinking into the environment, the robot can appropriately lean against the environment because the posture does not change even if agonist muscles relax.
  On the contrary, when sending the posture of leaving the environment, the robot can lean against the environment until the limit of $\Delta{\theta}_{max}$.
}%
{%
  テーブルに手を置いて休む動作の際の筋弛緩制御の働きについて考察する.
  重量物体を運ぶ際に, 一度テーブルの上に手を置いて身体を休め, また動作を再開する等の動作を想定している.
  \figref{figure:box-experiment}のように, 約5 kgの箱を抱え, 手をテーブルの上におろして身体を休める.
  この際, 認識やモデル誤差等から, 環境とロボットの間に内力が働きすぎず, 身体をテーブルに預ける動作が正確にできるとは限らない.
  本実験では, $S-p$と$E-p$のみを動作させ, 身体を休める際に$(\theta_{S-p}, \theta_{E-p})$がそれぞれ, 1: (-45, -80), 2: (-45, -85), 3: (-45, -90), 4: (-50, -90), 5: (-50, -95)になるように動作させた.
  これは, \figref{figure:box-tension}の下図のように, 休めたい身体部位とテーブルという環境の距離が1から5に進むについれて, めり込んだ状態から離れる状態へと移っていくような遷移である.

  筋弛緩制御を入れた場合と入れない場合における, 1から5の状態の$||\bm{T}_{current}||_{2}$を\figref{figure:box-tension}の上図に示す.
  筋弛緩制御を入れた場合は入れない場合に比べて, 無駄な拮抗筋による内力を抑制することができているため, より筋張力の値が小さくなっていることがわかる.
  また, 両者とも2の場合に最も筋張力が小さく, 1のように身体がテーブルにめり込むような姿勢を送ると筋張力が高まる.
  そして, 5のように身体が環境から離れていくと, 身体を環境に預けることができずに筋張力が増加していく.
  また, 筋弛緩制御を入れた場合と入れない場合における$||\bm{T}_{current}||_{2}$の差に着目する.
  その差は2や3のときに最も小さく, 1や5のときに増加する傾向にある.
  これは, 筋弛緩制御を入れない場合は身体を上手く環境に預けられるかどうかが敏感に変化するのに対して, 筋弛緩制御を入れることによって, 身体と環境の間の誤差を吸収して, 身体を上手く環境に預けることが可能となるからであると考えられる.
  身体が環境にめり込むような姿勢を送った場合には, 主動筋を弛緩させても身体姿勢に変化が起きないため, 最大限に身体を環境に預けることができる.
  逆に, 身体が環境から離れる方向に誤差がある場合には, $\Delta{\theta}_{max}$の分だけなるべく環境に身体を預けるような動作が生まれるのである.
}%

\subsection{Handle Operation}
\begin{figure}[t]
  \centering
  \includegraphics[width=0.9\columnwidth]{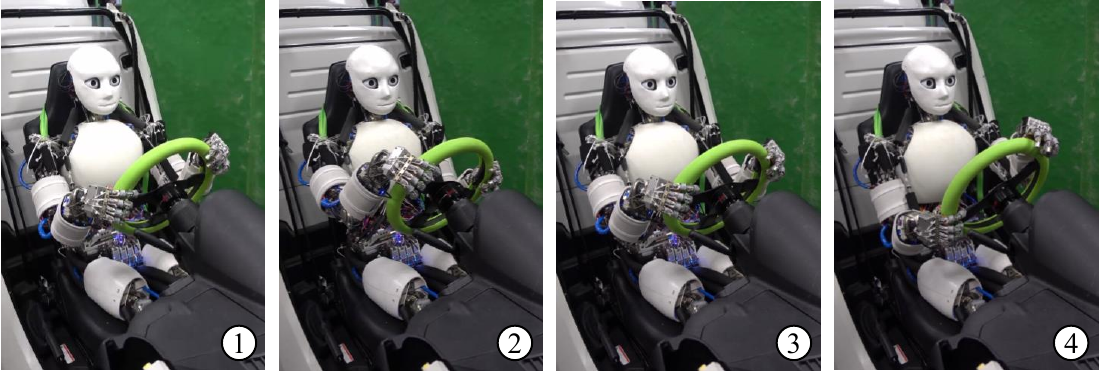}
  \caption{The experiment of handle operation.}
  \label{figure:handle-experiment}
\end{figure}

\begin{figure}[t]
  \centering
  \includegraphics[width=0.9\columnwidth]{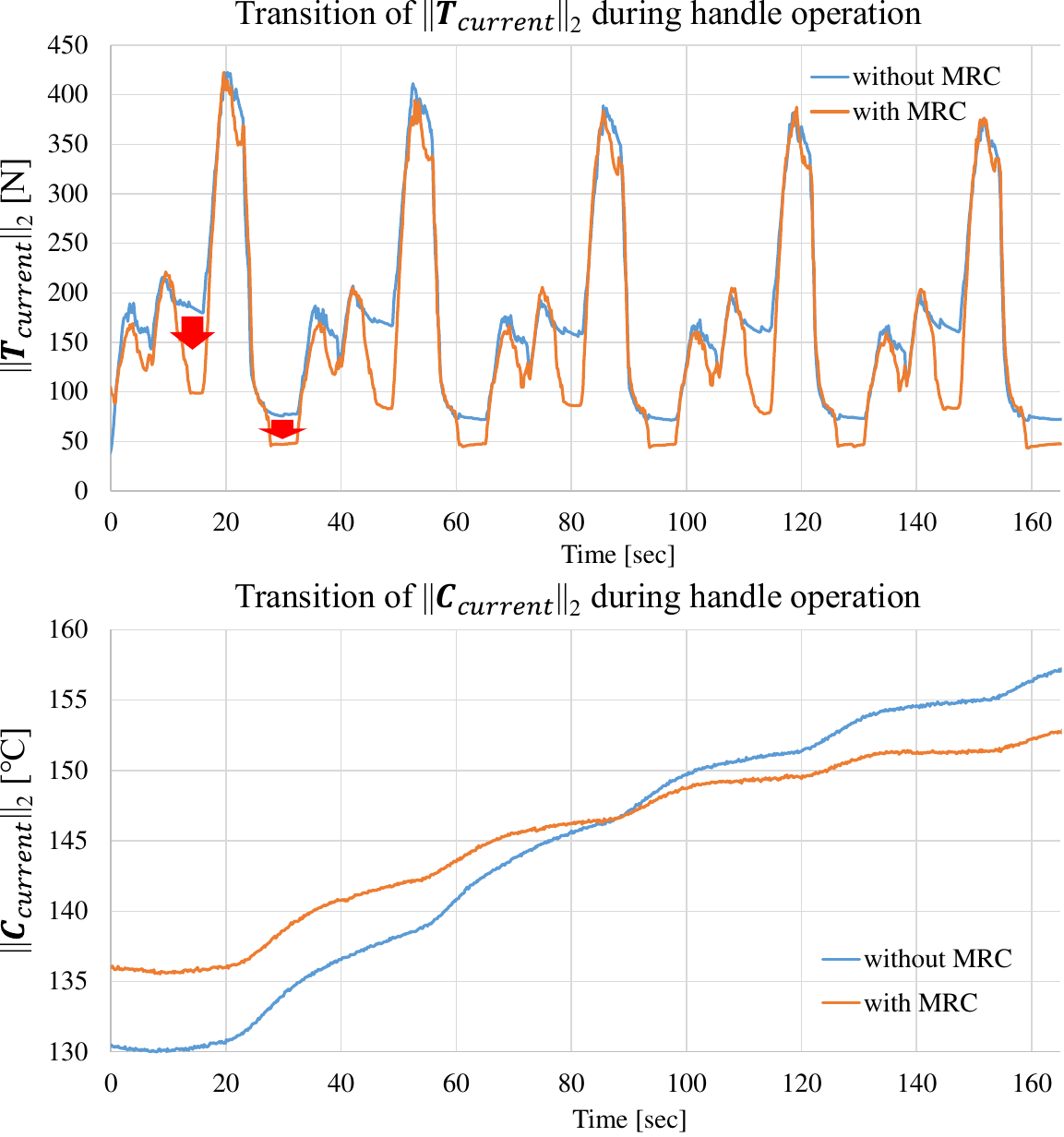}
  \caption{The transition of $||\bm{T}_{current}||_{2}$ and $||\bm{C}_{current}||_{2}$ during handle operation with and without muscle relaxation control.}
  \label{figure:handle-tension-thermo}
  \vspace{-3.0ex}
\end{figure}

\switchlanguage%
{%
  We examined the function of MRC when operating a handle.
  We solved Inverse Kinematics for the handle, and the robot rotated the handle to -45 and 45 deg with both arms 5 times as shown in \figref{figure:handle-experiment}.
  In this experiment, we sent the motion of rotating the handle 45 deg over 5sec, and the robot rested still for 5 sec at 0 deg and 3 sec at 45 or -45 deg.

  We show the transition of $||\bm{T}_{current}||_{2}$ and $||\bm{C}_{current}||_{2}$ with and without MRC in \figref{figure:handle-tension-thermo}.
  Compared to wihout MRC, muscle tension can be inhibited with MRC.
  Also, the muscle temperature is inhibited and the robot can operate the handle more continuously with MRC.

  In the handle operation, the handle and arms are constrained, and the robot can relax not only antagonist muscles but also agonist muscles.
  In the actual human handle operation, we are not continuously moving our arms but we are just holding the handle almost all the time.
  Similarly, MRC is effective for many motions including static states such as holding a train strap and typing a keyboard with the forearm on the desk.
}%
{%
  ハンドル操作の際の筋弛緩制御の働きについて考察する.
  ハンドルは環境に固定され, 回転方向のみの動きを持っている.
  ハンドルに対して逆運動学を解き, \figref{figure:handle-experiment}のように双腕で-45度から45度まで回転させる操作を5回行う.
  この際, 45度回転させる動作は5 secかけて行い, ハンドルが0 degのときは5 sec, 45または-45 degのときは3 sec静止する.

  筋弛緩制御を入れた場合と入れない場合における, $||\bm{T}_{current}||_{2}$の遷移を\figref{figure:handle-tension}に示す.
  筋弛緩制御を入れた場合は入れない場合に比べて, 筋張力を抑制することができていることがわかる.
  また, $||\bm{C}_{current}||_{2}$の遷移を\figref{figure:handle-thermo}に示す.
  筋弛緩制御を入れることで, 筋温度を抑制し, 継続的なハンドル操作を行うことが可能となる.

  ハンドル操作は常にハンドルと身体が拘束されるため, 拮抗筋だけでなく主動筋まで弛緩させることが可能である.
  実際の運転の際も常に身体を動かしているわけではなく, ほとんどの場合はハンドルに手をかけただけの状態である.
  このように, ハンドル操作をはじめとして, つり革を掴んだりキーボードを触るときに前腕を机に接触させたり, 多くの動作に置いて本制御が有効であると考える.
}%

\section{CONCLUSION} \label{sec:conclusion}
\switchlanguage%
{%
  In this study, we developed a muscle relaxation control (MRC) which inhibits useless internal muscle tension due to the error between the actual robot and its geometric model, and minimize necessary muscle tension by actively using the environment.
  Without changing the current posture, MRC relaxes muscles in order from less important ones and decreases internal muscle tension.
  Also, by constraining the body and environment, MRC can decrease the tension of agonist muscles.
  MRC can not only inhibit internal muscle tension during basic movements, but can also continuously enable tasks with environmental contact like the handle operation.

  It is important that a simple control method like MRC works well, and we can consider various reflex controls by slightly changing the behavior of each muscle in MRC.
  This concept may be applied to robots with pneumatic muscles, tensegrity robots, tendon-driven hands, etc, by changing the sensor states and control inputs that are handled.
  Also, human beings accumulate by how much to reflex muscles to move more smoothly than previously.
  In future works, we would like to combine simple reflex controls and learning control systems.
  Also, we would like to investigate the similarly and difference between this study and the behavior of human beings.
}%
{%
  本研究では, 筋骨格ヒューマノイドが幾何モデルとの誤差による無駄な筋内力を除去し, 環境を積極的に利用して必要な筋張力を最低限に保つ筋弛緩制御を開発した.
  現在のタスクに支障が出ない範囲内で, 重要度の低い筋から順に弛緩させていき, 筋内力を減らしていく.
  また, 環境によって身体を拘束することで, 重要度の高い主動筋の筋張力も抑えていくことができる.
  筋弛緩制御を用いることで, 通常の基本動作において内力の高まりを抑えられるだけでなく, ハンドル操作のような環境接触下におけるタスクも, 小さな力で遂行することが可能となった.

  本研究のようなシンプルな制御法が有効に働くことは重要であり, これを発展させて筋一本一本の挙動を変えていくことで, 様々な制御が可能となると考える.
  また, 人間の場合はどの程度弛緩させたかを記憶し, よりフォードフォワード的に身体動かしており, シンプルな反射型制御と学習型制御を統一的に扱うことについて今後考えていきたい.
}%

{
  \bibliographystyle{IEEEtran}
  \bibliography{main}

\begin{thebibliography}{10}
\providecommand{\url}[1]{#1}
\csname url@rmstyle\endcsname
\providecommand{\newblock}{\relax}
\providecommand{\bibinfo}[2]{#2}
\providecommand\BIBentrySTDinterwordspacing{\spaceskip=0pt\relax}
\providecommand\BIBentryALTinterwordstretchfactor{4}
\providecommand\BIBentryALTinterwordspacing{\spaceskip=\fontdimen2\font plus
\BIBentryALTinterwordstretchfactor\fontdimen3\font minus
  \fontdimen4\font\relax}
\providecommand\BIBforeignlanguage[2]{{%
\expandafter\ifx\csname l@#1\endcsname\relax
\typeout{** WARNING: IEEEtran.bst: No hyphenation pattern has been}%
\typeout{** loaded for the language `#1'. Using the pattern for}%
\typeout{** the default language instead.}%
\else
\language=\csname l@#1\endcsname
\fi
#2}}

\bibitem{nakanishi2013design}
Y.~Nakanishi, S.~Ohta, T.~Shirai, Y.~Asano, T.~Kozuki, Y.~Kakehashi,
  H.~Mizoguchi, T.~Kurotobi, Y.~Motegi, K.~Sasabuchi, J.~Urata, K.~Okada,
  I.~Mizuuchi, and M.~Inaba, ``{Design Approach of Biologically-Inspired
  Musculoskeletal Humanoids},'' \emph{International Journal of Advanced Robotic
  Systems}, vol.~10, no.~4, pp. 216--228, 2013.

\bibitem{wittmeier2013toward}
S.~Wittmeier, C.~Alessandro, N.~Bascarevic, K.~Dalamagkidis, D.~Devereux,
  A.~Diamond, M.~J{\"a}ntsch, K.~Jovanovic, R.~Knight, H.~G. Marques,
  P.~Milosavljevic, B.~Mitra, B.~Svetozarevic, V.~Potkonjak, R.~Pfeifer,
  A.~Knoll, and O.~Holland, ``{Toward Anthropomimetic Robotics: Development,
  Simulation, and Control of a Musculoskeletal Torso},'' \emph{Artificial
  Life}, vol.~19, no.~1, pp. 171--193, 2013.

\bibitem{jantsch2013anthrob}
M.~J{\"a}ntsch, S.~Wittmeier, K.~Dalamagkidis, A.~Panos, F.~Volkart, and
  A.~Knoll, ``{Anthrob - A Printed Anthropomimetic Robot},'' in
  \emph{Proceedings of the 2013 IEEE-RAS International Conference on Humanoid
  Robots}, 2013, pp. 342--347.

\bibitem{asano2016kengoro}
Y.~Asano, T.~Kozuki, S.~Ookubo, M.~Kawamura, S.~Nakashima, T.~Katayama,
  Y.~Iori, H.~Toshinori, K.~Kawaharazuka, S.~Makino, Y.~Kakiuchi, K.~Okada, and
  M.~Inaba, ``{Human Mimetic Musculoskeletal Humanoid Kengoro toward Real World
  Physically Interactive Actions},'' in \emph{Proceedings of the 2016 IEEE-RAS
  International Conference on Humanoid Robots}, 2016, pp. 876--883.

\bibitem{hirai1998asimo}
K.~Hirai, M.~Hirose, Y.~Haikawa, and T.~Takenaka, ``{The Development of Honda
  Humanoid Robot},'' in \emph{Proceedings of the 1998 IEEE International
  Conference on Robotics and Automation}, 1998, pp. 1321--1326.

\bibitem{hirukawa2004hrp}
H.~Hirukawa, F.~Kanehiro, K.~Kaneko, S.~Kajita, K.~Fujiwara, Y.~Kawai,
  F.~Tomita, S.~Hirai, K.~Tanie, T.~Isozumi, K.~Akachi, T.~Kawasaki, S.~Ota,
  K.~Yokoyama, H.~Handa, Y.~Fukase, J.~ichiro Maeda, Y.~Nakamura, S.~Tachi, and
  H.~Inoue, ``{Humanoid robotics platforms developed in HRP},'' \emph{Robotics
  and Autonomous Systems}, vol.~48, no.~4, pp. 165--175, 2004.

\bibitem{mizuuchi2006acquisition}
I.~Mizuuchi, Y.~Nakanishi, T.~Yoshikai, M.~Inaba, H.~Inoue, and O.~Khatib,
  ``{Body Information Acquisition System of Redundant Musculo-Skeletal
  Humanoid},'' in \emph{Experimental Robotics IX}, 2006, pp. 249--258.

\bibitem{kawaharazuka2018online}
K.~Kawaharazuka, S.~Makino, M.~Kawamura, Y.~Asano, K.~Okada, and M.~Inaba,
  ``{Online Learning of Joint-Muscle Mapping using Vision in Tendon-driven
  Musculoskeletal Humanoids},'' \emph{IEEE Robotics and Automation Letters},
  vol.~3, no.~2, pp. 772--779, 2018.

\bibitem{kawaharazuka2018bodyimage}
K.~Kawaharazuka, S.~Makino, M.~Kawamura, A.~Fujii, Y.~Asano, K.~Okada, and
  M.~Inaba, ``{Online Self-body Image Acquisition Considering Changes in Muscle
  Routes Caused by Softness of Body Tissue for Tendon-driven Musculoskeletal
  Humanoids},'' in \emph{Proceedings of the 2018 IEEE/RSJ International
  Conference on Intelligent Robots and Systems}, 2018, pp. 1711--1717.

\bibitem{kawaharazuka2019longtime}
K.~Kawaharazuka, K.~Tsuzuki, S.~Makino, M.~Onitsuka, Y.~Asano, K.~Okada,
  K.~Kawasaki, and M.~Inaba, ``{Long-time Self-body Image Acquisition and its
  Application to the Control of Musculoskeletal Structures},'' \emph{IEEE
  Robotics and Automation Letters}, vol.~4, no.~3, pp. 2965--2972, 2019.

\bibitem{asano2013loadsharing}
Y.~Asano, T.~Shirai, T.~Kozuki, Y.~Motegi, Y.~Nakanishi, K.~Okada, and
  M.~Inaba, ``{Motion Generation of Redundant Musculoskeletal Humanoid Based on
  Robot-Model Error Compensation by Muscle Load Sharing and Interactive Control
  Device},'' in \emph{Proceedings of the 2013 IEEE-RAS International Conference
  on Humanoid Robots}, 2013, pp. 336--341.

\bibitem{kawaharazuka2017antagonist}
K.~Kawaharazuka, M.~Kawamura, S.~Makino, Y.~Asano, K.~Okada, and M.~Inaba,
  ``{Antagonist Inhibition Control in Redundant Tendon-driven Structures Based
  on Human Reciprocal Innervation for Wide Range Limb Motion of Musculoskeletal
  Humanoids},'' \emph{IEEE Robotics and Automation Letters}, vol.~2, no.~4, pp.
  2119--2126, 2017.

\bibitem{osada2010addon}
M.~Osada, N.~Ito, Y.~Nakanishi, and M.~Inaba, ``{Realization of flexible motion
  by musculoskeletal humanoid ``Kojiro'' with add-on nonlinear spring units},''
  in \emph{Proceedings of the 2010 IEEE-RAS International Conference on
  Humanoid Robots}, 2010, pp. 174--179.

\bibitem{nakanishi2012absorption}
Y.~Nakanishi, T.~Izawa, T.~Kurotobi, J.~Urata, K.~Okada, and M.~Inaba,
  ``{Achievement of complex contact motion with environments by musculoskeletal
  humanoid using humanlike shock absorption strategy},'' in \emph{Proceedings
  of the 2012 IEEE/RSJ International Conference on Intelligent Robots and
  Systems}, 2012, pp. 1815--1820.

\bibitem{kawamura2016jointspace}
M.~Kawamura, S.~Ookubo, Y.~Asano, T.~Kozuki, K.~Okada, and M.~Inaba, ``{A
  Joint-Space Controller Based on Redundant Muscle Tension for Multiple DOF
  Joints in Musculoskeletal Humanoids},'' in \emph{Proceedings of the 2016
  IEEE-RAS International Conference on Humanoid Robots}, 2016, pp. 814--819.

\bibitem{shirai2011stiffness}
T.~Shirai, J.~Urata, Y.~Nakanishi, K.~Okada, and M.~Inaba, ``{Whole body
  adapting behavior with muscle level stiffness control of tendon-driven
  multijoint robot},'' in \emph{Proceedings of the 2011 IEEE International
  Conference on Robotics and Biomimetics}, 2011, pp. 2229--2234.

\bibitem{kawaharazuka2019musashi}
K.~Kawaharazuka, S.~Makino, K.~Tsuzuki, M.~Onitsuka, Y.~Nagamatsu, K.~Shinjo,
  T.~Makabe, Y.~Asano, K.~Okada, K.~Kawasaki, and M.~Inaba, ``{Component
  Modularized Design of Musculoskeletal Humanoid Platform Musashi to
  Investigate Learning Control Systems},'' in \emph{Proceedings of the 2019
  IEEE/RSJ International Conference on Intelligent Robots and Systems}, 2019,
  pp. 7294--7301.

\end{thebibliography}
}

\end{document}